\documentclass{article}
\usepackage{microtype}
\usepackage{graphicx}
\usepackage{subfigure}
\usepackage{booktabs} 

\usepackage{hyperref}
\usepackage[skip=4pt]{caption}

\usepackage[accepted]{icml2025}

\usepackage{multirow}
\usepackage{makecell}
\usepackage{natbib}
\usepackage{amsmath}
\usepackage{amssymb}
\usepackage{mathtools}
\usepackage{amsthm}

\usepackage[utf8]{inputenc} 
\usepackage[T1]{fontenc}    
\usepackage{hyperref}       
\usepackage{url}            
\usepackage{booktabs}       
\usepackage{amsfonts}       
\usepackage{nicefrac}       
\usepackage{microtype}      
\usepackage{xcolor}         
\usepackage{graphicx}
\usepackage{float}

\usepackage[capitalize,noabbrev]{cleveref}

\theoremstyle{plain}

\theoremstyle{definition}

\theoremstyle{remark}

\usepackage[textsize=tiny]{todonotes}

\icmltitlerunning{Evaluating Adversarial Protections for Diffusion Personalization: A Comprehensive Study}

\begin{document}

\pagestyle{empty}
\twocolumn[
\icmltitle{Evaluating Adversarial Protections for Diffusion Personalization: \\ A Comprehensive Study}



\icmlsetsymbol{equal}{*}

\begin{icmlauthorlist}
\icmlauthor{Kai Ye}{hdu}
\icmlauthor{Tianyi Chen}{ms}
\icmlauthor{Zhen Wang}{hdu}
\end{icmlauthorlist}

\icmlaffiliation{hdu}{School of Cyberspace, Hangzhou Dianzi University, Hangzhou, China}
\icmlaffiliation{ms}{Microsoft, Redmond, USA}

\icmlcorrespondingauthor{Zhen Wang}{wangzhen@hdu.edu.cn}



\icmlkeywords{Machine Learning, ICML}

\vskip 0.3in
]

\thispagestyle{empty}


\printAffiliationsAndNotice{}

\begin{abstract}
With the increasing adoption of diffusion models for image generation and personalization, concerns regarding privacy breaches and content misuse have become more pressing. In this study, we conduct a comprehensive comparison of eight perturbation-based protection methods—AdvDM, ASPL, FSGM, MetaCloak, Mist, PhotoGuard, SDS, and SimAC—across both portrait and artwork domains. These methods are evaluated under varying perturbation budgets, using a range of metrics to assess visual imperceptibility and protective efficacy. Our results offer practical guidance for method selection. Code is available at: https://github.com/vkeilo/DiffAdvPerturbationBench.
\end{abstract}
\section{Introduction}
\label{submission}

In recent years, generative models based on Denoising Diffusion Probabilistic Models (DDPMs)~\citep{ho2020denoising} have achieved remarkable progress in image synthesis. Unlike traditional Generative Adversarial Networks (GANs)~\citep{goodfellow2020generative}, which rely on adversarial training, diffusion models learn data distributions via forward and reverse sampling, enabling high-fidelity image generation. Building on this, researchers have introduced personalized generation techniques that let models quickly learn individual visual or artistic styles from only a few samples~\citep{ruiz2023dreambooth,kumari2023multi,gal2022image}, expanding the possibilities of customized content creation.

While personalized diffusion techniques offer creative potential, fine-tuning models with few samples raises significant privacy and copyright risks~\citep{liu2024disrupting}. Public or covert photos can be exploited to generate harmful content, and artists' styles can be misused without consent, underscoring the need for strong protections against privacy breaches and unauthorized style use. A promising defense involves adversarial perturbations, which disrupt model fine-tuning and hinder the reproduction of specific identities or styles~\citep{yang2021defending}.



To address the lack of standardized evaluation in current perturbation-based defenses for diffusion models, this paper proposes a unified benchmarking framework that systematically compares eight representative protection methods across two core tasks: identity protection and style imitation prevention. The framework supports multi-level perturbation control and incorporates a diverse set of perceptual and semantic metrics to jointly assess stealthiness and defensive efficacy. We apply this evaluation across two representative domains—portrait (VGGFace2) and artwork (WikiArt)—and report comprehensive experimental results that reveal trade-offs, robustness patterns, and sample-level variability. This study provides actionable insights for selecting appropriate protection strategies under different deployment constraints, and offers a generalizable foundation for future research on privacy-preserving generative models.

\section{Related Work}
\subsection{Diffusion Models and Customization}

Diffusion Models (DMs) generate high-quality images by gradually adding and removing noise, learning a reverse process to reconstruct samples~\citep{ho2020denoising}. Compared to VAEs~\citep{kingma2013auto} and GANs~\citep{goodfellow2020generative}, DMs offer superior image fidelity and training stability. Advances like classifier-free guidance~\citep{ho2022classifier} and Latent Diffusion Models (LDMs)~\citep{rombach2022high} further enhance expressiveness and efficiency, enabling scalable conditional generation.

To support personalization, methods like Textual Inversion~\citep{gal2022image} optimize embeddings without tuning model weights, while DreamBooth~\citep{ruiz2023dreambooth} fine-tunes diffusion models using a few reference images. Custom Diffusion~\citep{kumari2023multi} improves conditioning precision, and DiffuseKronA~\citep{marjit2024diffusekrona} reduces parameter overhead via Kronecker-based adapters. Despite these benefits, techniques raise privacy and copyright risks, as they may be exploited to generate harmful or plagiarized content, emphasizing the need for robust protection mechanisms.

\subsection{Perturbation-Based Privacy Protection}

Adversarial perturbation methods have emerged to protect users against unauthorized use of personal or artistic images. These approaches introduce small, carefully crafted noise to disrupt fine-tuning or generation.

AdvDM~\cite{liang2023adversarial} maximizes loss during denoising to hinder feature extraction. Anti-DreamBooth~\cite{van2023anti} combines surrogate models with learned perturbations to defend against DreamBooth-style personalization. PhotoGuard~\cite{salman2023raising} applies perturbations in the latent space, while GLAZE~\cite{shan2023glaze} targets style imitation with subtle visual noise. Mist~\cite{liang2023mist} introduces texture- and semantics-aware losses for robust protection. SimAC~\cite{wang2024simac} leverages frequency awareness and timestep selection to suppress identity features effectively.

MetaCloak~\cite{liu2024metacloak} uses meta-learning and surrogate models to ensure robust protection under diverse transformations. DisDiff~\cite{liu2024disrupting} disrupts prompt-based generation by erasing attention via cross-attention and schedule tuning.

Despite progress, existing methods often trade off stealthiness and protection strength, with performance varying across tasks and budgets. A unified evaluation framework is needed to compare these techniques systematically and guide real-world deployment.

\section{Method}

\subsection{Prerequisite Knowledge}

\textbf{Diffusion Models} generate high-quality data by denoising noisy samples through a learned reverse process. Starting from an original sample $\mathbf{x}_0$, the forward process adds Gaussian noise to form a sequence $\mathbf{x}_1, \mathbf{x}_2, \dots, \mathbf{x}_T$ via:
\begin{equation}
q(\mathbf{x}_t | \mathbf{x}_{t-1}) = \mathcal{N}(\mathbf{x}_t; \sqrt{1 - \beta_t} \mathbf{x}_{t-1}, \beta_t \mathbf{I})
\end{equation}
where $\beta_t$ controls the noise level. The reverse process $p_\theta(\mathbf{x}_{t-1} | \mathbf{x}_t)$ is learned by minimizing:
\begin{equation}
L_\text{denoise} = \mathbb{E}_{\mathbf{x}_0, t, \epsilon} \left[ \|\epsilon - \epsilon_\theta(\mathbf{x}_t, t)\|^2 \right]
\end{equation}
Here, $\epsilon$ is sampled noise, and $\epsilon_\theta(\cdot)$ is the predicted noise. Optimizing this loss enables faithful data generation.

However, when fine-tuned on few target samples, diffusion models can produce high-fidelity imitations, posing privacy and copyright risks. To counter this, adversarial perturbations $\delta$ can be added to inputs to hinder feature learning by disrupting the denoising objective during training.

\subsection{Perturbation Optimization Objective}

\begin{figure}[H]
    \centering
    \includegraphics[width=1\linewidth]{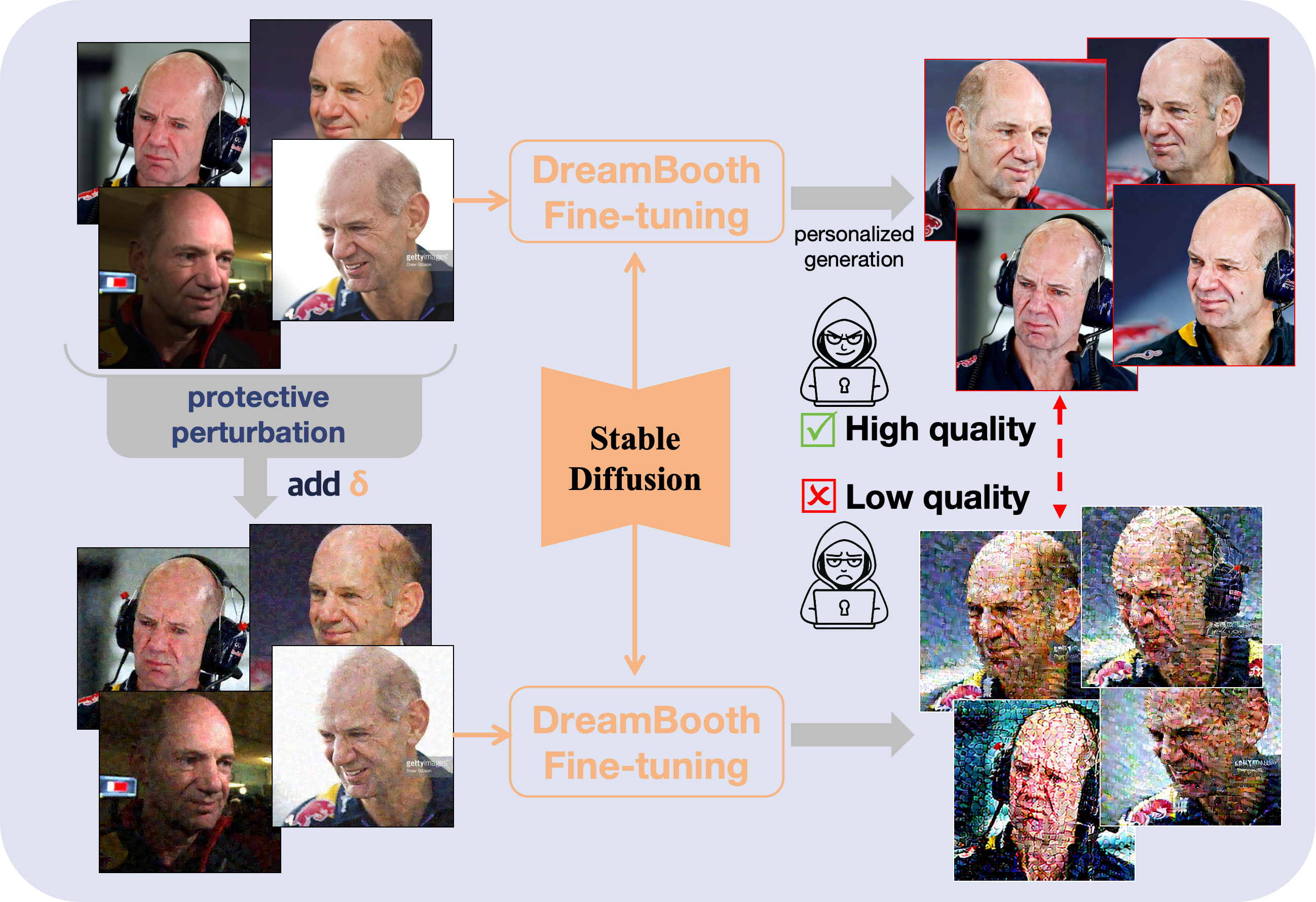}
    \caption{Overview of protective perturbation against diffusion-based personalization. Adding perturbation $\delta$ to training images degrades output quality after DreamBooth fine-tuning, preventing unauthorized identity replication.}
    \vspace{-1.7em}
    \label{fig:pipline}
\end{figure}

The goal of adversarial perturbations is to slightly modify input $\mathbf{x}$ to disrupt the generative process:
\begin{equation}
\delta^* = \arg\min_{\|\delta\| \leq \epsilon} \alpha L_\text{denoise}(\mathbf{x} + \delta)
\end{equation}
where $\delta^*$ is optimized via Projected Gradient Descent (PGD). This optimization hinders the model’s ability to learn accurate representations, reducing output fidelity. The overall protection pipeline is illustrated in Figure~\ref{fig:pipline}.

Different methods adopt distinct strategies for optimizing $\delta^*$ in this setting: \textbf{Mist} adds semantic-aware losses (e.g., texture loss), \textbf{SimAC} applies frequency-aware filtering and timestep control, while \textbf{MetaCloak} leverages surrogate feedback for robust adaptation under diverse conditions.

\subsection{Evaluation Framework Design}

To enable fair and consistent evaluation of protective perturbations for diffusion models, we introduce a unified framework that assesses both perturbation effectiveness and impact on customization. It supports batch processing across datasets and standardized metric-based comparisons, and consists of three modules:

\textbf{Perturbation Sample Generation}:  
We unify existing open-source implementations into a batch-compatible interface that supports custom datasets and standardized control over budgets and parameters.

\textbf{Customized Model Training}:  
Perturbed samples are used to train personalized models using a shared training pipeline to ensure consistent customization procedures.

\textbf{Sample Generation and Evaluation}:  
Customized models regenerate samples with consistent prompts, and evaluation uses diverse standardized metrics for comparability.

Our unified framework standardizes the evaluation of protective perturbations across tasks and metrics, enabling fair benchmarking of current methods and supporting future extensions. It reveals how performance varies with perturbation strength and task type, offering insights into the robustness and applicability of each method.

\setlength{\textfloatsep}{5pt plus 2pt minus 2pt}
\begin{figure}[H]
    \vspace{-0.5em}
    \centering
    \includegraphics[width=1\linewidth]{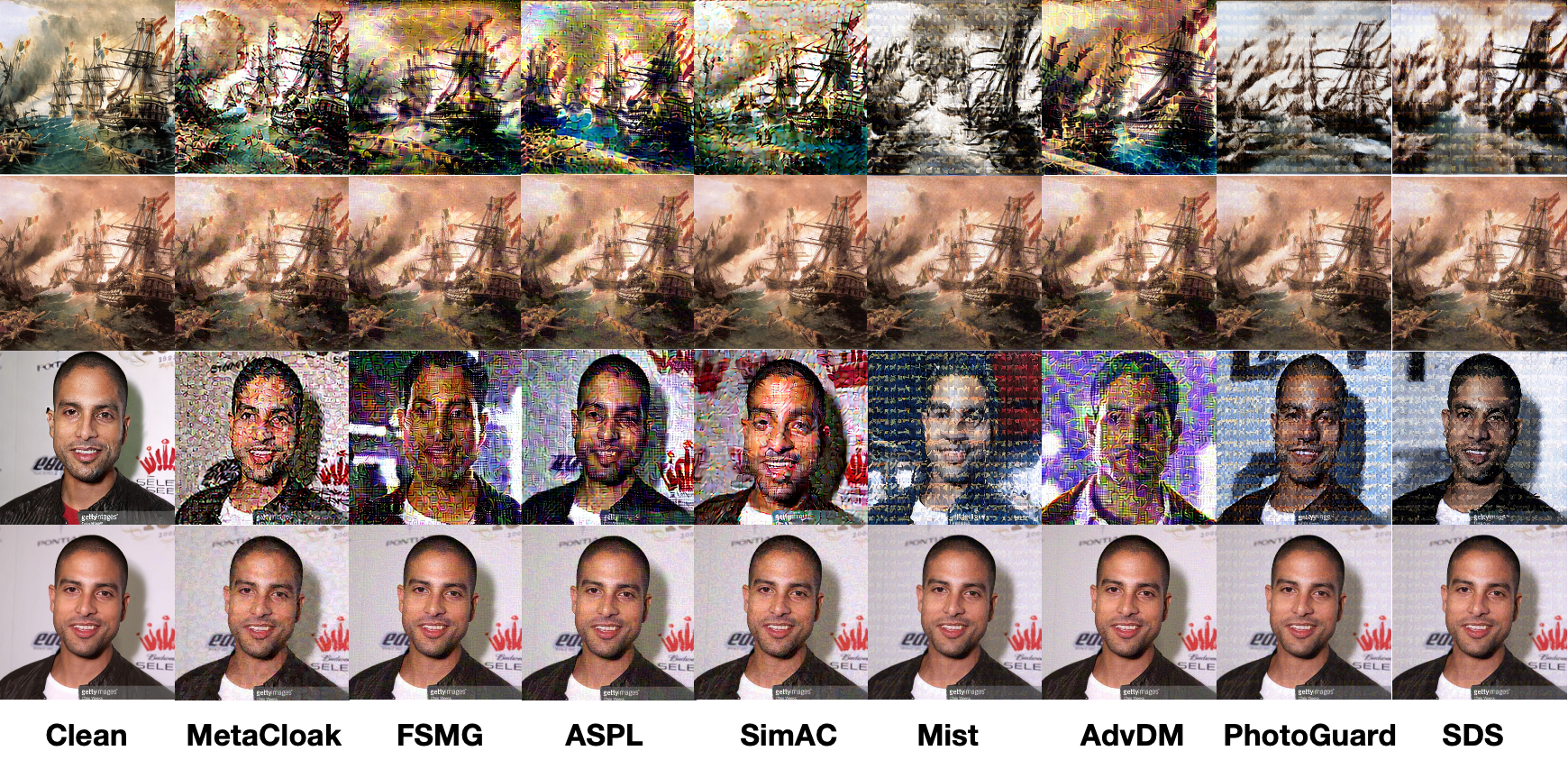}
    \caption{Comparison of perturbations and DreamBooth results on WikiArt (top) and VGGFace2 (bottom). Each column: perturbed input (bottom) and generation (top); clean sample on the left.}
    \label{fig:allinstance}
    \vspace{-1.7em}
\end{figure}
\section{Experiments}
\subsection{Selection of Datasets, Models, and Algorithms}
We select the \textbf{VGGFace2}~\citep{cao2018vggface2} and \textbf{WikiArt}~\citep{saleh2015large} datasets as our primary evaluation benchmarks. \textbf{VGGFace2} is a diverse face dataset with identity labels, commonly used for identity-preserving and privacy-related generation tasks. \textbf{WikiArt} contains artworks from various artists and styles, making it a standard benchmark for evaluating stylistic fidelity and cross-domain consistency.

For each dataset, we sample 50 homogeneous groups (same identity or artist), each with 8 images: 4 for clean reference and 4 for perturbation. Methods requiring pretraining are given 4 extra support images. All images are center-cropped and resized to 512×512.

We evaluated eight perturbation methods on a workstation equipped with 4 $\times$ RTX 4090 GPUs: ASPL~\citep{van2023anti}, FSGM~\citep{van2023anti}, SimAC~\citep{wang2024simac}, AdvDM~\citep{liang2023adversarial}, Mist~\citep{liang2023mist}, PhotoGuard~\citep{salman2023raising}, MetaCloak~\citep{liu2024metacloak}, and SDS~\citep{xue2023toward}. The step size is fixed at $1/255$ per iteration, with all other hyperparameters following default settings in their respective papers. See Figure~\ref{fig:allinstance} for visual examples of all methods.

\subsection{Evaluation Metrics}

We categorize the evaluation metrics into two major groups: \textbf{perturbation perceptibility metrics} and \textbf{generated image quality metrics}.See Appendix~A for detailed descriptions of each metric.



\textbf{Perturbation perceptibility} is evaluated using PSNR, LPIPS, SSIM, and CIEDE2000, which measure low-level visual differences and perceptual similarity.

\textbf{Generated image quality} is assessed via FID, PSNR, BRISQUE, LIQE, CLIP-IQA, and CLIP-IQAC, covering both perceptual fidelity and semantic consistency.

\subsection{Results and Discussion}
\subsubsection{Algorithm Performance Analysis}

\begin{table*}[t]
\centering
\caption{Comparison of perceptibility metrics across different datasets and perturbation budgets. Columns show results on the VGGFace2 datasets under average, low ($\epsilon=4/255$), and high ($\epsilon=16/255$) perturbation strengths. Higher PSNR and SSIM, and lower LPIPS and CIEDE2000 indicate better visual stealth.}
\vspace{6pt}
\label{tab:vis_metrics}
\resizebox{\textwidth}{!}{
\begin{tabular}{l|cccc|cccc|cccc}
\toprule
\multirow{2}{*}{\textbf{Method}} 
& \multicolumn{4}{c|}{\textbf{VGGFace2 Avg}} 
& \multicolumn{4}{c|}{\textbf{VGGFace2 Low ($\epsilon=4/255$)}} 
& \multicolumn{4}{c}{\textbf{VGGFace2 High ($\epsilon=16/255$)}} \\
\cmidrule(lr){2-5} \cmidrule(lr){6-9} \cmidrule(lr){10-13}
& PSNR $\uparrow$ & LPIPS $\downarrow$ & SSIM $\uparrow$ & CIEDE2000 $\downarrow$
& PSNR $\uparrow$ & LPIPS $\downarrow$ & SSIM $\uparrow$ & CIEDE2000 $\downarrow$
& PSNR $\uparrow$ & LPIPS $\downarrow$ & SSIM $\uparrow$ & CIEDE2000 $\downarrow$ \\
\midrule
AdvDM       & \textbf{33.141} & 0.274 & 0.785 & 2597.298   & 40.688 & 0.080 & 0.959 & 930.766   & 27.866 & 0.430 & 0.619 & 4132.730 \\
ASPL        & 31.821 & 0.311 & 0.757 & 2942.567   & 38.410 & 0.121 & 0.932 & 1263.338  & 27.137 & 0.463 & 0.605 & 4410.034 \\
FSGM        & 32.751 & 0.289 & 0.797 & 2479.031   & 38.596 & 0.120 & 0.934 & 1224.997  & \textbf{28.975} & 0.412 & \textbf{0.692} & \textbf{3413.226} \\
MetaCloak   & 33.119 & 0.274 & \textbf{0.799} & \textbf{2444.655}   & \textbf{41.460} & \textbf{0.064} & \textbf{0.968} & \textbf{862.729} & 27.451 & 0.444 & 0.640 & 3880.745 \\
Mist        & 32.716 & 0.264 & 0.773 & 2511.432   & 40.314 & 0.067 & 0.955 & 866.704   & 27.363 & 0.430 & 0.603 & 4084.828 \\
PhotoGuard  & 32.553 & 0.266 & 0.767 & 2546.989   & 40.142 & 0.068 & 0.953 & 882.197   & 27.224 & 0.433 & 0.595 & 4128.659 \\
SDS         & 32.547 & 0.275 & 0.776 & 2501.789   & 40.144 & 0.071 & 0.953 & 885.339   & 27.243 & 0.447 & 0.619 & 3960.694 \\
SimAC       & 32.667 & \textbf{0.255} & 0.787 & 2570.826   & 39.246 & 0.089 & 0.943 & 1123.767  & 28.116 & \textbf{0.389} & 0.650 & 3774.565 \\
\bottomrule
\end{tabular}
}
\vspace{-1em}
\end{table*}

Table~\ref{tab:vis_metrics} reports perceptibility scores on VGGFace2 under overall, low, and high perturbation levels. \textbf{MetaCloak} excels in pixel-level and structural similarity for identity protection, while \textbf{FSGM} performs best for artistic styles. \textbf{SimAC} consistently yields the lowest LPIPS, indicating strong deep-feature stealth. Stealthiness varies with budget: \textbf{MetaCloak} leads at $4/255$, but \textbf{FSGM} and \textbf{SimAC} excel at $16/255$, stressing the importance of budget-aware method selection.

Protective perturbations aim to degrade personalized outputs post-finetuning by disrupting input semantics. Table~\ref{tab:gen_quality} shows generation quality on VGGFace2 and WikiArt. On VGGFace2, \textbf{MetaCloak} yields the highest FID, and \textbf{SimAC} the lowest CLIP-IQAC, indicating strong disruption to structure and semantics. \textbf{Mist}, \textbf{PhotoGuard}, and \textbf{SDS} lead on distortion-based metrics (BRISQUE, PSNR), reflecting greater visual degradation. \textbf{SimAC} also scores well in LIQE, underscoring its broad disruptive impact.

Across both datasets, method effectiveness shows consistent patterns: \textbf{SimAC} and \textbf{MetaCloak} excel in semantic disruption, while \textbf{PhotoGuard}, \textbf{SDS}, and \textbf{Mist} are stronger in low-level distortion. This highlights the need to align method choice with protection goals—semantic vs. visual degradation. Similar to perceptibility trends, performance varies with perturbation level: \textbf{SimAC} dominates at $4/255$, whereas \textbf{MetaCloak} leads at $16/255$ on most quality metrics. Notably, \textbf{FSGM} retains high generation quality despite good stealth at high budgets, suggesting limited disruption. Detailed results are provided in Appendix~B.

\begin{table*}[t]
\small
\setlength{\tabcolsep}{1.5pt}
\centering
\caption{Comparison of generation quality metrics across VGGFace2 and WikiArt datasets. Strong protection corresponds to higher BRISQUE and FID, and lower LIQE, CLIP-based scores, and PSNR.}
\label{tab:gen_quality}
\begin{tabular}{l|cccccc|cccccc}
\toprule
\multirow{2}{*}{\textbf{Method}} 
& \multicolumn{6}{c|}{\textbf{VGGFace2}} 
& \multicolumn{6}{c}{\textbf{WikiArt}} \\
\cmidrule(lr){2-7} \cmidrule(lr){8-13}
& LIQE $\downarrow$ & BRISQUE $\uparrow$ & FID $\uparrow$ & PSNR $\downarrow$ & CLIPIQA $\downarrow$ & \makecell{CLIP\\IQAC} $\downarrow$ 
& LIQE $\downarrow$ & BRISQUE $\uparrow$ & FID $\uparrow$ & PSNR $\downarrow$ & CLIPIQA $\downarrow$ & \makecell{CLIP\\IQAC} $\downarrow$ \\
\midrule
clean        & 2.922 & 9.810  & 199.949 & 9.181 & 0.860 & 0.269  & 2.069     & 19.989      & 320.613       & 10.017     & 0.721     & 0.231      \\
AdvDM        & 1.041 & 3.221  & 344.216 & 8.521 & 0.623 & -0.323 & 1.105 & 9.665  & 379.464 & 9.312 & 0.542 & -0.251 \\
ASPL         & \textbf{1.015} & 6.247  & 356.409 & 8.561 & 0.562 & \textbf{-0.411} & \textbf{1.065} & 9.798  & 393.336 & \textbf{9.223} & 0.487 & -0.350 \\
FSGM         & 1.030 & -3.199 & 328.851 & 8.472 & 0.635 & -0.312 & 1.068 & 6.613  & 379.033 & 9.228 & 0.529 & -0.283 \\
MetaCloak    & 1.191 & 10.462 & \textbf{381.347} & 8.587 & 0.597 & -0.373 & 1.382 & 18.120 & 395.469 & 9.952 & 0.465 & -0.348 \\
Mist         & 1.756 & 16.898 & 379.146 & 8.801 & 0.762 & -0.159 & 1.845 & 23.552 & 410.557 & 10.006& 0.489 & -0.279 \\
PhotoGuard   & 1.986 & \textbf{19.858} & 362.558 & 8.885 & 0.767 & -0.112 & 2.095 & \textbf{24.982} & 405.753 & 9.747 & 0.515 & -0.243 \\
SDS          & 1.806 & 17.968 & 374.678 & 8.916 & 0.784 & -0.136 & 1.888 & 23.952 & \textbf{413.541} & 9.620 & 0.535 & -0.258 \\
SimAC        & 1.048 & 12.779 & 362.534 & \textbf{8.217} & \textbf{0.526} & -0.406 & 1.134 & 14.626 & 409.888 & 9.421 & \textbf{0.415} & \textbf{-0.409} \\
\bottomrule
\end{tabular}
\vspace{-1em}
\end{table*}

\subsubsection{Analysis of Sample-Level Variability}

\begin{figure}[H]
    \centering
    \includegraphics[width=1\linewidth]{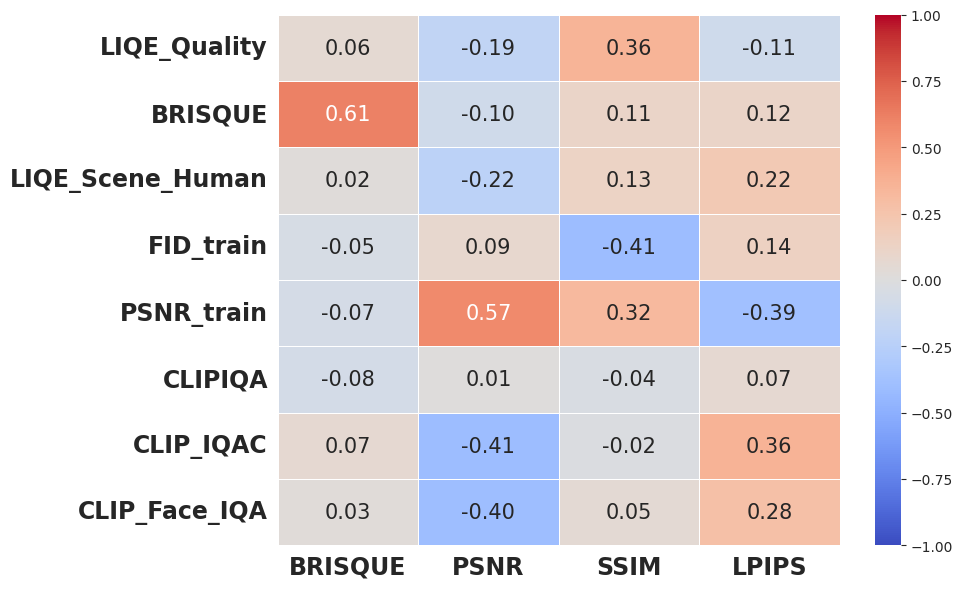}
    \caption{Pearson correlations between output quality (rows) and training image quality over 50 clean VGGFace2 identities.}
    \label{fig:corr}
\end{figure}

In our experiments, we found that the performance of perturbation methods fluctuates significantly across individual identity samples. This variability is largely due to the strong sensitivity of downstream personalization models (e.g., DreamBooth) to training image properties, rather than the inherent robustness of the perturbation methods themselves. Correlation analysis (Figure~\ref{fig:corr}) reveals that higher consistency among training images improves output PSNR, LIQE, and FID—indicating enhanced structural fidelity and stable identity learning. Additionally, a strong BRISQUE correlation ($r = 0.615$) suggests that output quality is closely tied to the quality of the input samples.

However, excessive consistency may harm semantic generalization. We observed a significant negative correlation between CLIP-IQAC and training consistency (e.g., $r = -0.41$ with PSNR), suggesting that overly redundant samples can cause overfitting to local features, leading to semantic drift. To mitigate this, we recommend a “structurally mixed training set” strategy—combining both consistent and diverse subsets in each identity’s training set. This hybrid design balances structural fidelity and semantic robustness, offering practical guidance for data selection in perturbation-based protection.


\section{Conclusion}

With the increasing use of diffusion models for personalized image generation, protecting user privacy has become a critical and urgent concern. This paper presents a unified evaluation framework and systematically compares eight representative perturbation methods across two key dimensions: perceptibility and generation quality. Experimental results show that no single method dominates across all metrics or perturbation budgets. For instance, \textbf{SimAC} excels in perceptual stealth at low budgets, while \textbf{MetaCloak} proves more effective at degrading output quality under stronger perturbations, emphasizing the need to match strategies to specific deployment constraints.

Further analysis reveals that the effectiveness of protection methods varies significantly across individual samples, largely due to the sensitivity of personalization models to training image structure. Correlation studies on VGGFace2 suggest that training image consistency improves output fidelity but may hinder semantic generalization when overly excessive. We propose a “structurally mixed training set” strategy to better balance these factors. Experiments on WikiArt confirm consistent performance trends across domains, supporting the general applicability of these proposed methods. Future work will extend the framework to broader tasks and explore scalable privacy-preserving techniques for multimodal diffusion systems.

\clearpage
\bibliography{example_paper}
\bibliographystyle{icml2025}



\clearpage
\appendix
\section{Metric Descriptions}

To ensure a comprehensive and fair evaluation of perturbation-based protection methods, we adopt a diverse set of metrics that reflect both perceptual quality and feature-level fidelity. These metrics are divided into two main categories:

\textbf{Perturbation Perceptibility Metrics}:

\begin{itemize}
    \item \textbf{PSNR (Peak Signal-to-Noise Ratio)}: Measures the pixel-level similarity between the original and perturbed images. Higher PSNR indicates lower distortion.
    \item \textbf{SSIM (Structural Similarity Index)}~\citep{wang2004image}: Evaluates structural similarity between two images, emphasizing luminance, contrast, and structure. Higher values indicate better visual similarity.
    \item \textbf{LPIPS (Learned Perceptual Image Patch Similarity)}~\citep{zhang2018unreasonable}: A deep feature-based distance that correlates well with human perception. Lower scores indicate better perceptual similarity.
    \item \textbf{CIEDE2000}~\citep{sharma2005ciede2000}: A perceptual color difference metric based on the CIE Lab color swpace. In our evaluation, we compute the L2 norm across all per-pixel CIEDE2000 values to obtain a global color distortion measure. Lower values indicate higher perceptual similarity in color.
\end{itemize}

\textbf{Generated Image Quality Metrics}:

\begin{itemize}
    \item \textbf{BRISQUE (Blind/Referenceless Image Spatial Quality Evaluator)}~\citep{mittal2012no}: A no-reference quality score based on natural scene statistics. Lower values denote higher perceived quality.
    \item \textbf{LIQE (Learning-based Image Quality Evaluator)}~\citep{zhang2023blind}: A learned quality model trained on subjective ratings to predict perceptual quality. Lower scores are better.
    \item \textbf{FID (Fréchet Inception Distance)}~\citep{heusel2017gans}: Measures the distributional distance between generated and reference images using deep features. Lower values reflect better alignment with the target domain.
    \item \textbf{CLIP-IQA}~\citep{wang2023exploring}: Computes perceptual similarity using CLIP features, capturing high-level alignment between generated and reference content. Lower values indicate greater degradation.
    \item \textbf{CLIP-IQAC (CLIP Image-Query Alignment Consistency)}~\citep{liu2024metacloak}: Evaluates the consistency between a prompt (text) and generated images via CLIP similarity. Lower scores imply more effective protection.
\end{itemize}

\clearpage
\onecolumn
\section{Additional Evaluation Metrics}
\subsection{Generation Quality Results at Varying Perturbation Strengths}

\begin{table}[h]
\footnotesize
\centering
\caption{Generation quality metrics of each method under different perturbation strengths $r \in \{4, 8, 12, 16\}/255$, on VGGFace2}
\label{tab:appendix_gen_quality}
\makebox[\textwidth][c]{
\begin{tabular}{c l cccccccc}
\toprule
$r$ & Method & LIQE\_Quality & BRISQUE & \makecell{LIQE\\Scene\\Human} & FID & PSNR & CLIPIQA & CLIP-IQAC & \makecell{CLIP\\Face\\IQA} \\
\midrule

$0/255$ & Clean Avg   & 1.646 & 8.838  & 0.810 & 283.106 & 8.760 & 0.725 & -0.109 & 0.279 \\
\midrule

\multirow{8}{*}{$4/255$}
& AdvDM       & 1.149 & 4.929  & 0.918 & 269.781 & 8.712 & 0.712 & -0.105 & 0.256 \\
& ASPL        & 1.052 & 6.066  & 0.900 & 289.036 & 8.791 & 0.652 & -0.252 & 0.258 \\
& FSGM        & 1.097 & -0.691 & 0.936 & 279.614 & 8.559 & 0.702 & -0.193 & 0.267 \\
& MetaCloak   & 1.499 & 9.157  & 0.849 & 274.512 & 8.741 & 0.725 & -0.116 & 0.206 \\
& Mist        & 2.386 & 10.538 & 0.714 & 284.934 & 8.855 & 0.805 & 0.026  & 0.357 \\
& PhotoGuard  & 2.430 & 13.953 & 0.746 & 265.942 & 8.961 & 0.797 & 0.058  & 0.369 \\
& SDS         & 2.399 & 13.002 & 0.751 & 277.158 & 8.875 & 0.814 & 0.037  & 0.342 \\
& SimAC       & 1.154 & 13.747 & 0.666 & 323.874 & 8.584 & 0.593 & -0.329 & 0.179 \\
\midrule

\multirow{8}{*}{$8/255$}
& AdvDM       & 1.011 & 2.183  & 0.864 & 330.384 & 8.526 & 0.631 & -0.316 & 0.253 \\
& ASPL        & 1.004 & 2.827  & 0.876 & 339.068 & 8.552 & 0.566 & -0.409 & 0.242 \\
& FSGM        & 1.015 & -4.390 & 0.923 & 312.659 & 8.427 & 0.632 & -0.300 & 0.279 \\
& MetaCloak   & 1.153 & 11.021 & 0.491 & 372.615 & 8.523 & 0.639 & -0.385 & 0.225 \\
& Mist        & 1.642 & 17.696 & 0.389 & 376.225 & 8.977 & 0.768 & -0.191 & 0.357 \\
& PhotoGuard  & 1.934 & 19.669 & 0.413 & 354.655 & 8.826 & 0.787 & -0.115 & 0.394 \\
& SDS         & 1.647 & 18.096 & 0.489 & 367.428 & 8.906 & 0.787 & -0.152 & 0.380 \\
& SimAC       & 1.020 & 13.588 & 0.581 & 356.480 & 8.295 & 0.535 & -0.414 & 0.187 \\
\midrule

\multirow{8}{*}{$12/255$}
& AdvDM       & 1.003 & 3.555  & 0.820 & 373.292 & 8.486 & 0.586 & -0.400 & 0.236 \\
& ASPL        & 1.001 & 7.378  & 0.814 & 380.787 & 8.392 & 0.532 & -0.476 & 0.227 \\
& FSGM        & 1.004 & -4.248 & 0.884 & 360.779 & 8.476 & 0.619 & -0.364 & 0.276 \\
& MetaCloak   & 1.079 & 11.466 & 0.286 & 429.366 & 8.543 & 0.546 & -0.466 & 0.186 \\
& Mist        & 1.495 & 18.461 & 0.248 & 416.673 & 8.616 & 0.759 & -0.217 & 0.359 \\
& PhotoGuard  & 1.768 & 22.404 & 0.350 & 407.268 & 8.861 & 0.746 & -0.186 & 0.374 \\
& SDS         & 1.583 & 19.586 & 0.313 & 411.580 & 9.044 & 0.789 & -0.175 & 0.355 \\
& SimAC       & 1.015 & 12.368 & 0.601 & 376.901 & 8.057 & 0.505 & -0.439 & 0.180 \\
\midrule

\multirow{8}{*}{$16/255$}
& AdvDM       & 1.001 & 2.219  & 0.769 & 403.408 & 8.359 & 0.563 & -0.469 & 0.211 \\
& ASPL        & 1.001 & 8.716  & 0.811 & 416.747 & 8.507 & 0.496 & -0.507 & 0.209 \\
& FSGM        & 1.003 & -3.465 & 0.881 & 362.353 & 8.424 & 0.586 & -0.394 & 0.265 \\
& MetaCloak   & 1.032 & 10.204 & 0.226 & 448.894 & 8.542 & 0.477 & -0.523 & 0.097 \\
& Mist        & 1.503 & 20.896 & 0.205 & 438.753 & 8.757 & 0.718 & -0.255 & 0.330 \\
& PhotoGuard  & 1.812 & 23.406 & 0.278 & 422.369 & 8.891 & 0.739 & -0.204 & 0.342 \\
& SDS         & 1.595 & 21.188 & 0.271 & 442.548 & 8.838 & 0.745 & -0.253 & 0.309 \\
& SimAC       & 1.005 & 11.414 & 0.538 & 392.880 & 7.931 & 0.470 & -0.443 & 0.156 \\
\bottomrule
\end{tabular}
}
\end{table}
\clearpage
\twocolumn
\end{document}